%% file: lexnorm-journal.tex
\documentclass[smallextended]{svjour3}
\usepackage{times,latexsym}
\usepackage{url}
\usepackage[T1]{fontenc}
\usepackage{amsmath}
\usepackage{natbib}
\usepackage{array,multirow,graphicx}

\usepackage{float}
\restylefloat{table}

\usepackage{graphicx}
\usepackage{subfig}
\usepackage{xcolor}

\usepackage{xspace,mfirstuc,tabulary}

\usepackage{amsmath}

\title{Word-level Lexical Normalisation using Context-Dependent Embeddings}

\author{Michael~Stewart,
        Wei~Liu,
        and~Rachel~Cardell-Oliver}

\institute{Michael Stewart, Wei Liu, and Rachel Cardell-Oliver \at
              The University of Western Australia \\
              \email{michael.stewart@research.uwa.edu.au, \{wei.liu, rachel.cardell-oliver\}@uwa.edu.au}           
}

\newcolumntype{C}[1]{>{\centering\let\newline\\\arraybackslash\hspace{0pt}}m{#1}}

\begin{document}

\definecolor{darkline}{rgb}{0.82, 0.82, 0.82}
\definecolor{darklinetext}{rgb}{0.3, 0.3, 0.3}
\definecolor{unsurecolor}{rgb}{0.5, 0.5, 0.5}
\definecolor{lightgray}{rgb}{0.8, 0.8, 0.8}

\newcommand{\filler}[1]{{
	{\color{darklinetext} \textit{#1}}
	\color{darkline} \newline
	\rule{\columnwidth}{6pt} \newline
	\rule{\columnwidth}{6pt} \newline
	\rule{\columnwidth}{6pt} \newline
	\rule{\columnwidth}{6pt} \newline
	\rule{\columnwidth}{6pt} \newline
	\rule{\columnwidth}{6pt} \newline
	\rule{\columnwidth}{6pt} \newline
	\rule{\columnwidth}{6pt} \newline
	\rule{\columnwidth}{6pt} \newline
	\rule{\columnwidth}{6pt} \newline
	\rule{\columnwidth}{6pt} \newline
}}

\newcommand{\absfiller}[1]{{
	{\color{darklinetext} \textit{#1}}
	\color{darkline} \newline
	\rule{0.75\columnwidth}{6pt} \newline
	\rule{0.75\columnwidth}{6pt} \newline
	\rule{0.75\columnwidth}{6pt} \newline
	\rule{0.75\columnwidth}{6pt} \newline
	\rule{0.75\columnwidth}{6pt} \newline
	\rule{0.75\columnwidth}{6pt} \newline
	\rule{0.75\columnwidth}{6pt} \newline
	\rule{0.75\columnwidth}{6pt} \newline
	\rule{0.75\columnwidth}{6pt} \newline
	\rule{0.75\columnwidth}{6pt} \newline
	\rule{0.75\columnwidth}{6pt} \newline
	\rule{0.75\columnwidth}{6pt} \newline
	\rule{0.75\columnwidth}{6pt} \newline
	\rule{0.75\columnwidth}{6pt} \newline
	\rule{0.75\columnwidth}{6pt} \newline
}}

\maketitle
\begin{abstract}

Lexical normalisation (LN) is the process of correcting each word in a dataset to its canonical form so that it may be more easily and more accurately analysed. Most lexical normalisation systems operate at the character-level, while word-level models are seldom used. Recent language models offer solutions to the drawbacks of word-level LN models, yet, to the best of our knowledge, no research has investigated their effectiveness on LN. In this paper we introduce a word-level GRU-based LN model and investigate the effectiveness of recent embedding techniques on word-level LN. Our results show that our GRU-based word-level model produces greater results than character-level models, and outperforms existing deep-learning based LN techniques on Twitter data. We also find that randomly-initialised embeddings are capable of outperforming pre-trained embedding models in certain scenarios. Finally, we release a substantial lexical normalisation dataset to the community. 

\end{abstract}


\input{section-introduction}
\input{section-related-work}
\input{section-algorithm}

\input{section-datasets}

\input{section-experiments}

\input{section-results}

\input{section-conclusion}


\section{Acknowledgements} This research was funded by an Australian Postgraduate Award Scholarship and a UWA Safety Net Top-up Scholarship.

\bibliographystyle{spbasic.bst}
\bibliography{references}

\end{document}

%% file: section-introduction.tex
\section{Introduction}

Lexical normalisation (LN) is the process of restoring ill-formed tokens in a dataset to their canonical form~\citep{han2011lexical}. It is an indesposable pre-processing critical to almost all natural language processing tasks such as entity recognition, sentiment analysis, text classification, and automatic question answering~\citep{hua2015short}. 

The primary focus of LN research through the shared task is on short text data. Short text is common across many domains, the most prevalent of which is social media. Another exemplary short text domain is corporate operational reports, such as workplace incident reports, maintenance work orders, daily supply chain delay logs. For example, for workplace incident logs, reporters keep track of incidents that have occurred by typing a short description to log the incident. For example:

\texttt{ee was standing next to komatsu and notied twinge in l shoulder}

This log demonstrates a multitude of errors typical to domain specific short text: \texttt{ee} is an abbreviation for \texttt{employee}, \texttt{notied} is a spelling error, and \texttt{l} is an abbreviation for \texttt{left}. The fact that \texttt{komatsu} is a non-English but correct term in the context of mining equipment makes the task even more challenging. One cannot easily rely on an English dictionary to normalise tokens, but domain specific out of vocabulary (OOV) words are by far the most difficult to deal with.

These accident reports are extremely valuable for decision making if domain knowledge can be discovered from these collections of textual logs. For example, from mining a set of workplace accident logs from 2013, we have discovered that serious injuries related to haul trucks tend to be back and neck injuries. Thesse discoveries are extremely valuable for workplace safety protection. Analysing social media data also provides a wide variety of insights, such as the ability to determine public opinion and predict election results~\citep{anstead2014social}. Without LN, however, it is impossible to perform such analysis with accuracy. 

Most LN approaches operate at the character level. While character-level models allow for vastly reduced memory usage over word-level models, they have the tendency to predict slightly-incorrect character sequences, resulting in reduced accuracy~\citep{leeman2015ncsu_sas_sam}. Word-level models, which rely heavily on word embeddings, are seldom used for LN because misspelled words do not have corresponding word embedding vectors unless the embedding model is trained upon the training corpus itself, which is typically much smaller than the corpora used for embedding training, and thus do not capture enough semantic meaning.

There have recently been great advances in the area of character and word representation. Unlike the previous state-of-the-art embedding models, FastText~\citep{bojanowski2017enriching} is trained on character n-grams, as opposed to words, allowing for the generation of word vectors for words that do not appear in the training corpus. ELMo~\citep{peters2018elmo} is trained at the character-level and provides the facility to generate vectors that are dependent on context, allowing different word vectors for tokens with the same surface form. BERT~\citep{devlin2018bert} uses a bidirectional transformer and has improved the state-of-the-art in a wide range of NLP tasks.

The effectiveness of these state-of-the-art embedding models have not yet been evaluated on the lexical normalisation of short text data. In addition, almost all LN research focuses specifically on social media, practical normalisation techniques for domain specific short text domains such as industrial accident reports are yet to be developed.

In light of this gap in research, we investigate the effectiveness of recent embedding models for LN on three different short text domains. We introduce a word-level LN model and demonstrate the strong performance of word-level models when utilising the latest embedding models. We also show that randomly-initiated embeddings can outperform pre-trained embedding models when the dataset contains many sparse, lexically-similar terms such as measurements. Finally, we release a substantial manually-annotated lexical normalisation dataset. We hope that this dataset proves useful for lexical normalisation research and inspires other researchers to implement their own models.

%% file: section-related-work.tex
\section{Related Work}


Lexical normalisation (LN) is a relatively recently-proposed problem in natural language processing. Prior to the conception of the term by \citet{han2011lexical}, similar research targeted spell checking, which only focuses on the correction of spelling errors in natural language.  \citet{kukich1992techniques} introduce a number of early techniques, including n-gram analysis and dictionary lookup. \citet{wong2008enhanced} introduce a comprehensive scoring system that takes multiple factors into account when normalising a term, including edit distance, the presence of the term in an abbreviation dictionary, and the left and right context words. More recently, \citet{whitelaw2009using} incorporated data from the World Wide Web into an n-gram based model, significantly improving spell checking accuracy using unsupervised learning.



Early lexical normalisation techniques typically make use of similarity-based measures to select the most likely candidate for each erroneous token. \citet{han2011lexical} utilise a linear kernal SVM classifier to identify ill-formed tokens and normalise them using morphophonemic similarity. This research was later extended with a wider variety of experimentation, including the introduction of a word-level classification model~\citep{han2013lexical}. \citet{ahmed2015lexical} investigate the effectiveness of phonetic matching algorithms on social media data, namely Refined Soundex and Peter Norvig's Algorithm. 

Many recent lexical normalisation techniques are deep learning based, negating the need for handcrafted features. \citet{cox2015ncsu_sas_sam} introduce a deep encoding pipeline-based system that features two separate feed-forward networks, a flagger and normaliser. The \textit{flagger} component, a simple network with two output nodes, determines whether a token is ill-formed. The \textit{normaliser} component handles normalisation, predicting the most likely correct character for each character in the input sequence. Any tokens that do not appear in a pre-built dictionary are normalised to the most similar token in the dictionary using the ancillary \textit{conformer} module.

\citet{mott2015ncsu_sas_wookhee}'s LSTM-based model predicts the edit distance operations required to transform the input token into the output token. The model normalises tokens at the character-level, taking a fixed window of words as input along with the POS tag of each token in the window. This allows the model to incorporate contextual information into its predictions, in contrast to the deep encoding model which normalises tokens purely based on their character sequences.

Text-to-speech (TTS) is another area related to lexical normalisation that has recently been the focus of the Google Text Normalisation Challenge~\citep{sproat2016rnn}. This challenge saw the introduction of a variety of RNN-based models. The first model uses two separate LSTMs: one to identify the \textit{channel} (the possible normalisations of a token), and another to model the language. The \textit{channel model} predicts the most likely correct form of the given word, while the \textit{language model} aims to predict the next word given the current word in a similar manner to Word2Vec~\citep{mikolov2013distributed}. The outputs of the two LSTMs are combined at decoding time to predict the output token. The paper also  introduces an attention-based RNN model, inspired by~\citet{chan2016listen}. The model normalises fixed windows of words at the character-level. The network learns to label each window with the output tag (i.e. the correct form) of the centre word.

The deep models built to handle TTS tasks require large amounts of training data to achieve such high performance, and do not necessarily work well on the much smaller amounts of data available in the short text domain. \citet{sproat2016rnn}'s best performing model was trained on a dataset of over 1.1 billion English words and 290 million Russian words, which is on a completely different scale than any available short text dataset.

Despite the popularity of LN on social media data and TTS, other short text domains receive little attention.~\citet{stewart2018lexical} introduce a pipeline for LN on short text industrial data that specifically targets the unique types of errors prevalent in short text data, namely acronyms, spelling errors, and domain-specific terms. A bidirectional LSTM is trained to classify each OOV token into a particular error type, and each token is then normalised according to its error type. Despite its high performance on short text data, it performs poorly on acronym expansion and relies heavily on a dictionary building approach. The dictionary builder automatically normalises any tokens in the evaluation dataset that always map to the same token in the training dataset.


%% file: section-algorithm.tex
\section{Model Architecture}

Hereby we propose a word-level lexical normalisation model uses a bidirectional gated recurrent unit (GRU)~\citep{cho2014learning}. The GRU can be considered a simpler alternative to the LSTM. With only two gates, GRUs converge quicker and require fewer parameter updates than the LSTM~\citep{chung2014empirical}. The Bi-GRU, as opposed to a unidirectional GRU, allows for both forward and backwards contexts to be taken into account when predicting the correct form of the word. This is necessary because the correct form of a word, such as ``l'' (whose correct form may be ``left''), can be determined by the words that precede it (``he turned l'') or follow it (``l leg was injured''). It can often depend on both preceding and following words as well (``he injured his l leg''). Combining a forward and backwards GRU to form a Bi-GRU allows for the model to effectively utilise this information.

\begin{figure}[!ht]
    \begin{center}
        \includegraphics[width=\columnwidth]{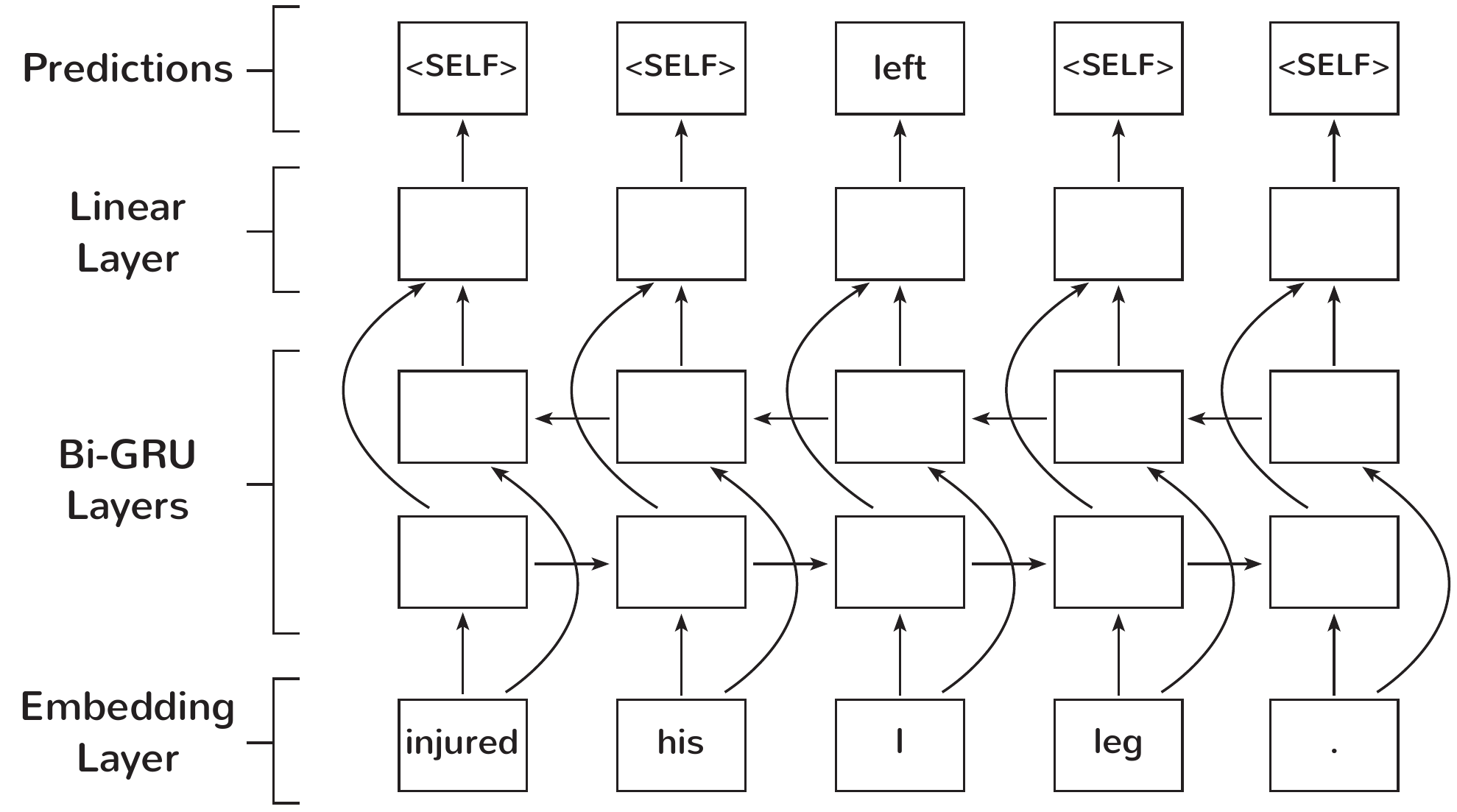}
        \caption{The architecture of the word-level lexical normalisation model.}
        \label{fig:wordlevelmodel}
    \end{center}
\end{figure}

Our model, as shown in Figure~\ref{fig:wordlevelmodel}, takes a sequence of tokens as input and predicts a sequence of labels as output. Each input token is labelled with exactly one output token corresponding to the correct form of that token. In a similar manner to~\citep{sproat2016rnn}, a special \texttt{<SELF>} label is used to denote words that are already in their correct form. This helps to significantly speed up training by reducing the number of possible labels to predict, while also improving performance by reducing the number of infrequently-occurring labels. In Figure~\ref{fig:wordlevelmodel}, the original shorthand token ``l'' is normalised to ``left'' while all other tokens in the sentence do not require normalisation and are hence labelled as \texttt{<SELF>}.

Our model may also be used as a character-level model. Instead of predicting the correct form of each word in a sentence, it can instead predict the correct character corresponding to each character in a word. In Section~\ref{subsec:wordvschar} we evaluate the effectiveness of our model when applied at the character-level.


\subsection{Embedding layer initialisation}

The embedding layer of the word-level model may be initialised using a number of embedding techniques, including \textit{pre-trained word embeddings}, \textit{cooccurence-based embeddings}, and \textit{random initialisation} using a probability distribution. The embedding layer plays a significant role in the lexical normalisation model. Word-level embeddings, especially context-sensitive embeddings, are highly effective in NLP tasks such as question answering and semantic entailment where the meaning of a word is important. We argue that lexical normalisation, on the other hand, should rely more on embeddings that have been trained at the character or character n-gram level. We therefore carry out a comprehensive study on word embedding methods.

\subsubsection{Word embedding models}

One of the most popular word embedding models is Word2Vec~\citep{mikolov2013distributed}, which contains two separate training models: \textit{Skip-gram}, whereby the model attempts to predict the surrounding context of a word, and \textit{Continuous bag of words (CBOW)}, where the model predicts a word given a surrounding context. The skip-gram model aims to maximise the log likelihood across the predictions of all words in the context window of size $c$ surrounding word $w$:

\begin{equation}
    \frac{1}{T} \sum^{T}_{t = 1} \sum_{-c \leq j \leq c, j \neq 0} \log p(w_{t+j} | w_t) 
\end{equation}

Word2Vec operates at the word-level, meaning the characters that comprise each word are irrelevant to the generated embeddings. Ignoring the morphology of words is a major limitation to Word2Vec, particularly for lexical normalisation whereby many words in the training corpus are misspelled.

FastText~\citep{bojanowski2017enriching} operates similarly to the skip-gram variant of Word2Vec but instead represents each word as a bag of character ngrams alongside the word itself. The predictions made by the model are evaluated using a modified scoring function:

\begin{equation}
    s(w, c) = \sum_{g \in \mathcal{G}_w} \mathbf{z}_g^\top \mathbf{v}_c
\end{equation}

where $w$ is a word, $\mathcal{G}_w$ is the set of ngrams appearing in $w$, G is the size of each ngram, and $\mathbf{z}_g$ is the vector representation of each n-gram $g$.

FastText's word embeddings are based upon the sum of the embeddings of each ngram in $w$, making it amenable to lexical normalisation where misspelled words and correctly-spelled words share similar ngrams. 

ELMo~\citep{peters2018elmo} constructs embeddings using learned functions of a bidirectional language model (Bi-LM). A Bi-LM features two separate recurrent LSTMs. The forward model computes the probability of a context window of tokens $(t_1, t_2, ..., t_N)$ based on the probability of each token appearing $t$ given a sequence of previous tokens:

\begin{equation}
    p(t_1, t_2, ..., t_N) = \prod^{N}_{k=1} p(t_k | t_{k+1}, t_{k+2}, ..., t_N)
\end{equation}

The backward model is identical to the forward model, but operates in reverse, predicting the previous token given a sequence of future tokens. ELMo combines these two models and jointly maximises the log likelihood of both directions.

Unlike FastText, ELMo is trained purely at the character level, and every word embedding obtained through the bidirectional language model is context-dependent. This means that the embedding vector of words appearing in many different contexts depend on the sentence the word appears in as well as the characters that comprise the word itself. This is highly advantageous for lexical normalisation considering the number of abbreviations, acronyms, and spelling errors that can normalise to a multitude of different terms depending on their context.

BERT~\citep{devlin2018bert} is based upon the encoder stack of the bidirectional transformer model~\citep{vaswani2017attention}, an encoder-decoder model structure that combines feed-forward layers with a multi-headed attention mechanism. In contrast to ELMo, BERT learns deep bidirectional representations by performing a procedure known as the ``masked language model''. For every input sentence to the model some number of terms are masked at random, and the model learns to predict the original terms.

The primary distinction between BERT and ELMo with respect to LN is that ELMo is trained at the character level whereas BERT is trained on wordpieces (fragments of words). BERT also offers two pre-trained models, BERT\textsubscript{BASE} and BERT\textsubscript{LARGE}, which may be finetuned on domain-specific datasets. 

\subsubsection{Co-occurrence models}
\label{subsec:coocex}
A more simplistic and quick way to initialise the embedding layer is to use a co-occurrence model. Given a set of words $V$ and set of documents $D$, one can generate a co-occurrence matrix $X$ of size $|V| \times |D|$ which denotes the documents that each word appears in. Three common approaches to populating this matrix are as follows:

\begin{enumerate}
    \item One-hot: When a word $w$ appears in a document $d$, $X_{w, d} = 1$;
    \item Cumulative: $X_{w, d}$ denotes how many times $w$ appears in $d$; and
    \item TF-IDF: $X_{w, d}$ denotes the Term Frequency - Inverse Document Frequency of $w$ with respect to $d$.
\end{enumerate}

We explore the effectiveness of these various methods in Section~\ref{subsec:cooc}.

\subsubsection{Probability distributions}

Another way to initialise the weights of the embedding layer is to generate a probability distribution for the embedding vector of each word. This is seldom practised due to the high performance gain from using word embedding models in a wide variety of NLP tasks. Many standard deep learning libraries initialise the weights using a uniform distribution by default, though other distributions such as normal and cauchy may be used instead.

\subsection{Bi-GRU layers}

The forward and backwards Bi-GRU layers serve to encode the embedded inputs and provide the final output layer with a representation that may be mapped to a sequence of label probabilities. The update gate $z$ and reset gate of the GRU $r$ are calculated as follows, where $U$ and $W$ are weight matrices to be learned, $x_t$ is the input at time $t$, and $h_t$ is the hidden state at time $t$~\citep{koehn2009statistical}:

\begin{subequations}
\begin{align}
z_t = \sigma(x_{t}U^z + h_{t-1}W^z + b_z) \\
r_t = \sigma(x_{t}U^r + h_{t-1}W^r + b_r)
\end{align}
\end{subequations}

The hidden state of the GRU is then calculated as follows:

\begin{subequations}
\begin{align}
\widetilde{h}_t = tanh(x_{t}U^h + (r_t \odot h_{t-1})W^h) \\
h_t = (1- z_t) * h_{t-1} + z_t * \widetilde{h}_t
\end{align}
\label{eqn:gru}
\end{subequations}

The input sequence is first sent to the embedding layer and then passes through two GRU layers, each of which encodes the input sequence from left-to-right and right-to-left.

\subsection{Output layer}

The output of the GRU is translated back to the label space by a linear layer:

\begin{equation}
y' = h_tA^{\top} + b
\label{eqn:linear}
\end{equation}

where $h_t$ is the hidden state of the final timestep of the GRU and $A$ is a weight matrix that is learned.

The log softmax operation is then performed on each row of the prediction matrix $y'$  in order to produce a probability distribution over all possible labels:

\begin{equation}
y'_i={log(\frac {e^{y'_{i}}}{\sum _{j=1}^{J}e^{y'_{j}}})} \\
\end{equation}

where $J$ is the number of possible labels (i.e. the number of columns in $y'$).

The inputs to our model are padded with a special ``padding token'' so that every input is the same length. In order to avoid this special token significantly affecting the loss, a mask is applied to the output vectors prior to performing the loss function. This function effectively removes all rows in the output vectors corresponding to the indexes of the padding tokens in $y$. Given $Y$ is a vector of the indexes of the correct class labels corresponding to predictions $y'$, and that the padding token always has an index of 0, each row and column in the prediction matrix $y'$ is updated as follows:

\begin{subequations}
\begin{align}
    f(x) := \begin{cases}0,&x=0\\{}1,&x\neq 0\end{cases} \\
    y'_{i, j} = f(Y_{i, j}) \times y'_{i, j}
\end{align}
\label{eqn:masking}
\end{subequations}

Cross entropy is then used to calculate the loss:

\begin{equation}
loss = -\frac{\sum_{c=1}^J y_{c, o} \times y'_{c, o}}{n} 
\label{eqn:loss}
\end{equation}

where $y$ is a binary indicator determining whether the class label $c$ is correct for observation $o$, $J$ is the number of possible labels, and $n$ is the number of non-masked tokens in the prediction.








%% file: section-datasets.tex
\section{Datasets}

\subsection{US Accidents dataset}

This paper contributes a new short text lexical normalisation dataset\footnote{The dataset is available at https://github.com/Michael-Stewart-Webdev/us-accidents-dataset.} to be made publicly available for LN research. It is an annotated version of the US Accident Injuries dataset\footnote{Data.gov. \\ \textit{https://catalog.data.gov/dataset/accident-injuries}}, a US government collection of information on accidents reported by mine operators and contractors. 

Our US Accidents dataset contains 10,000 accident reports, randomly split into a training set of 9,000 reports and an evaluation set of 1,000 reports. These 10,000 reports were randomly sampled from the original Accident Injuries dataset, which contains approximately 228,000 documents in the form of a CSV file. 

\subsubsection{Dataset annotation}

\begin{figure}[!ht]
    \begin{center}
        \includegraphics[width=\columnwidth]{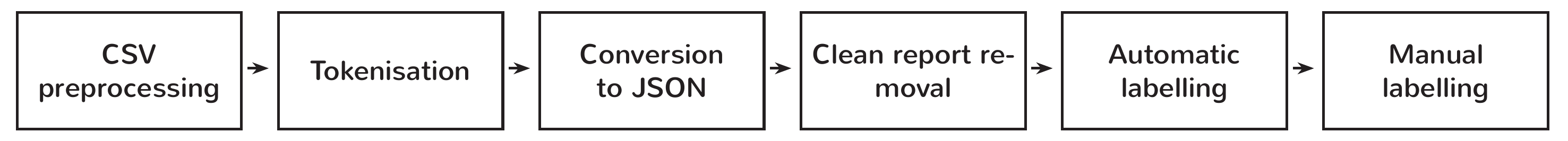}
        \caption{The dataset creation process pipeline.}
        \label{fig:pipeline}
    \end{center}
\end{figure}

Figure~\ref{fig:pipeline} shows a pipeline of the dataset creation process. In order to create the dataset, we first removed all columns from the CSV file aside from the index and textual description columns. The textual descriptions were then tokenized using NLTK\footnote{NLTK. \textit{http://www.nltk.org}} and converted to JSON format along with the corresponding row indices. We then removed reports that did not contain at least one non-English word, ensuring that the dataset does not contain any documents that are not useful for training. 

The dataset was curated using an annotation tool built specifically for this paper\footnote{Source code will be released upon publication.}. Part of the process was automated, namely the use of a dictionary and domain regex rules to automatically label very commonly-occurring terms such as ``ee" (``employee") and ``approx" ("approximately"). Every document was read and annotated in order to remove any false positives arising from this automated approach. The majority of the annotation had to be completed manually due to the wide variety of errors present in the dataset. 

During annotation, we removed all documents containing terms we did not recognise (540 in total). We continued the annotation process until we had annotated exactly 10,000 documents.

Care was taken to ensure consistency across the labelling of domain-specific terms, particularly with regards to terms that were commonly written with slight variations. For example, the term ``cross-cut'' (a type of underground passageway) is written in a multitude of different ways in the original dataset: ``xc'', ``x-c'', ``crosscut'', ``cross cut'', and so on.  In order to ensure that each variant was labelled with the exact same token, we performed a separate, automated normalisation process after completing the manual annotation. During annotation, we maintained a list of terms that were found to be written many different ways. We then used a script to search the labels of each document in the annotated dataset and perform substitutions for these terms using regular expressions.

\subsubsection{Dataset evaluation}

We evaluate our dataset using a script that determines the type of each token, as shown in Tables~\ref{tab:us-dataset-table-ne} and ~\ref{tab:us-dataset-table-e}. This categorisation is based upon the input and output token, as well as the following input and output token in the case of ``split words'' (words that reporters have accidentally separated with a space character). A token is considered erroneous if the input token does not match the corresponding output token.

The most common error category in the dataset is abbreviations, primarily due to the abundance of ``ee'' being used to denote ``employee''. Spelling errors comprise approximately one quarter of all errors. In addition to the aforementioned ``split words'', the dataset also contains many ``joined words'' whereby the report authors have omitted space characters between words.  One of the more challenging aspects of the dataset is the prevalence of misspelled domain-specific terms, such as ``x-c'' being used to denote ``cross-cut'', and measurements such as ``8'h'' denoting ``8' high''. There are also numerous unnecessary tokens, such as the common use of ``.'' characters to denote abbreviations.


\begin{table}[t]
\centering
\begin{tabular}{l c c}

\textbf{Token type}   &     \textbf{\# Train}     &  \textbf{\# Test}  \\ \hline
English words                  & 305,684 & 33,713 \\
Punctuation                    & 35,224 & 3,914 \\
Dates/numbers                  & 11,708 & 1,288 \\
Domain-specific terms          & 3,806 & 428 \\
\hline
\textbf{Total} & 356,422 & 39,343 \\
\hline

\end{tabular}
\caption{\label{tab:us-dataset-table-ne} The number of non-erroneous tokens in the training and evaluation set, categorised by token type.
  }
\end{table}


\begin{table}[t]
\centering
\begin{tabular}{l c c}

\textbf{Token type}   &     \textbf{\# Train}     &  \textbf{\# Test}  \\ \hline
Abbreviations                  & 10,539 & 1,106 \\
Spelling errors                & 4,898 & 529 \\
Joined words                   & 2,129 & 258 \\
Split words\textsuperscript{*} & 1,659 & 186 \\
DST spelling errors            & 1,374 & 176 \\
Unnecessary tokens             & 922 & 108 \\
Acronyms                       & 866 & 88 \\
\hline
\textbf{Total} & 22,387 & 2,451 \\
\hline

\end{tabular}
\caption{\label{tab:us-dataset-table-e} The number of erroneous tokens in the training and evaluation set, categorised by token type. \\ \textsuperscript{*}Since split words always comprise two tokens their count in the table is actually double the number of occurrences.
  }
\end{table}




\subsection{Australian Accidents dataset}

The Australian Accidents dataset, first described by~\citep{stewart2018lexical}, features 2,322 accident reports from the Australian mining industry. There are 159,560 tokens in total, split into a training and testing set of 90\% and 10\% of the documents respectively. These reports were split into sentences in order to reduce the sequence lengths. While it is similar to the US Accident dataset in terms of the domain-specific language used, the Australian dataset contains fewer split word, joined words, and unnecessary tokens. The reports are also written differently. The American dataset has more emphasis on the inclusion of excavation measurements and locations, while the Australian dataset is written in a more general manner. 

\subsection{Twitter dataset}

For bench marking purposes we also use the Twitter dataset, the same dataset used in WNUT 2015~\citep{baldwin2015shared}. It features 4,917 documents (2,950 train, 1,967 test) with 73,806 tokens in total. It is distinct from the two datasets mentioned above, primarily with regards to the number of "at mentions", "hashtags" and "url" tokens. It is significantly noisier than the two other datasets as a result of these tokens, as well as a wider variety of misspellings and non-English general domain terms.

%% file: section-experiments.tex
\section{Experiments}


\subsection{Model parameters}

The Bi-GRU and feed-forward models were trained with a batch size of 80 and a learning rate of 0.1. The models were optimised using momentum-based SGD ($\beta = 0.9$). The Bi-GRU models were trained with 2 GRU layers of 512 neurons each, with 50\% dropout. The character-level Bi-GRU models had a character embedding dimension of 100.

\subsection{Data pre-processing}


The special tokens in the Twitter dataset (hashtags, at-mentions and URLs) were removed from the training set prior to the training of the character-level model, as those tokens never require normalisation. Any tokens that were completely non-alphabetical were removed from the training sets of the Australian Accidents and Twitter datasets.



\subsection{Word embedding training}

To evaluate our word-level model we generate embeddings using four word embedding models: Word2Vec~\citep{mikolov2013distributed}, FastText~\citep{bojanowski2017enriching}, ELMo~\citep{peters2018elmo}, and BERT~\citep{devlin2018bert}. The embedding vectors from all embedding models except BERT are used as input to our Bi-GRU word-level model. The BERT embeddings are fed through a single feed-forward layer followed by a softmax classifier, as recommended by the BERT paper.

 The Word2Vec and FastText models were trained with an embedding dimension of 512 and a window size of 7. ELMo was also trained with an embedding dimension of 512. For BERT, as recommended in the BERT paper, we pre-trained BERT\textsubscript{BASE} on each of our own corpora. To extract the BERT embeddings we took the final hidden state of the transformer (embedding dimension 768). After testing the results of various iteration limits we settled on 50 epochs for Word2Vec, FastText, and ELMo, and ~5 epochs for BERT. Despite ELMo being trained for 10 times as many epochs as BERT, both models took approximately the same amount of time to complete training.

The US Accidents embedding models were trained on the full ~228,000-document US Accidents corpus. The Australian Accidents models were trained on the small 2,322-document Australian Accident dataset. The Australian Accidents embedding models were initially trained on the US Accidents corpus (with all US spelling converted to UK English, combined with the Australian corpus), but it was found to produce inferior results to the Australian corpus alone. The Twitter models were trained on a 1.6 million-document sentiment analysis dataset from Kaggle\footnote{\textit{Kaggle}. \\ https://www.kaggle.com/kazanova/sentiment140}, which was combined with the Twitter dataset used here for evaluation. Training ELMo and BERT on this Twitter dataset took approximately one week each on a Titan X GPU.





\subsection{Post-processing}
\label{subsec:postprocessing}
To evaluate the best performing model overall, we performed two separate post-processing steps after normalisation. The first technique, \textit{dictionary normalisation}, automatically normalises tokens in the test data when those tokens always map to the same label in the training data~\citep{mott2015ncsu_sas_wookhee}. The token ``ee'', for example, always normalises to ``employee'' in the US Accident dataset, so any predictions made by our deep learning models for ``ee'' are automatically replaced by ``employee''.

The \textit{flagger} is a simple character-level neural network that determines whether a token should or should not be normalised. For our evaluation we adapt \citep{cox2015ncsu_sas_sam}'s flagger module to use a GRU architecture instead of a feed-forward architecture in an effort to improve accuracy. Rather than the final layer containing two neurons, our GRU-based flagger model takes the last timestep of the GRU and passes it through the linear layer to map the predictions back to the label space.

%% file: section-results.tex
\section{Results}

Our results investigate the following:

\begin{enumerate}
	\item \textbf{Word embedding evaluation}: Which embedding technique produces the most accurate word-level lexical normalisation (LN) results, and why?
    \item \textbf{Word-level vs character-level lexical normalisation}: How does word-level LN compare to character-level LN?
    \item \textbf{Performance comparison with existing systems}: How well does our top performing model compare to existing social media LN systems?
\end{enumerate}

\subsection{Word embedding evaluation}

\label{subsec:wordembeval}

Our first result criterion aims to determine the most effective word embedding technique for word-level lexical normalisation, and why it is the most effective. We investigate and compare three types of embedding techniques.

In a similar manner to ~\citep{sproat2016rnn}, we augment each dataset with a \texttt{<SELF>} tag to denote words that do not require normalisation. Unless explicity stated (as in Section~\ref{subsubsec:selfeffectiveness}), the models here are trained on this augmented dataset.

\subsubsection{Probability distributions for randomised embeddings}
\label{subsec:bestdist}

\begin{table}[!ht]
\centering
\begin{tabular}{c | c  | c | c | c}


        & \textbf{Uniform} & \textbf{Normal}    & \textbf{Cauchy} & \textbf{Best} \\ \hline
        
\multicolumn{5}{c}{Weights learned} \\
\hline
Twitter & 0.7946          & \textbf{0.8022} & 0.7896 & 0.8022\\
Aus Acc & 0.7950 &  \textbf{0.7983} & 0.7899  & 0.7983 \\
US Acc  & 0.8312          & 0.8316 & \textbf{0.8368} & 0.8368\\
\hline
\multicolumn{5}{c}{Weights frozen} \\
\hline
Twitter & 0.7946 & \textbf{0.8000}  & 0.7862          & 0.8000\\ 
Aus Acc     & 0.7950 & \textbf{0.7983}  & 0.7899         & 0.7983 \\ 
US Acc  & 0.8300  & 0.8321  &  \textbf{0.8378} & 0.8378 \\ 

\end{tabular}
\caption{\label{tab:distributions-table} The F1 scores of our word-level GRU on each of the three datasets when the embedding layer is initialised with one of three probability distributions.
  }
\end{table}

We begin our investigation into word-level lexical normalisation by first determining the best performing probability distribution for the initialisation of embedding layer weights.

To perform this task we initialise the weights of the embedding layer of our word-level GRU with three commonly-used probability distributions:  Uniform distribution in the range $\{-2~...~2\}$, Normal distribution ($\mu = 0$, $\sigma = 1.0$), and Cauchy distribution ($\mu = 0$, $\sigma = 1.0$). We then train the model on each dataset twice: once with non-frozen weights, and once with frozen weights. The models' performance is then evaluated on the corresponding test set.

Figure~\ref{tab:distributions-table} shows the corresponding F1 scores of each probability distribution. There is little variance between distributions, although the model trained on the US Accidents dataset benefited relatively significantly from the cauchy distribution. Freezing the weights appears to make no difference to the F1-score. Overall we find that the most appropriate probability distribution to initialise the weights of the embedding layer depends entirely on the dataset.

\subsubsection{Cooccurrence-based embeddings}
\label{subsec:cooc}

\begin{table}[!ht]
\centering
\begin{tabular}{c | c  | c | c | c}

        & \textbf{One-hot} & \textbf{Cumul.}    & \textbf{TF-IDF} & Best \\ \hline
\multicolumn{5}{c}{No PCA} \\
\hline
Twitter & 0.6557 & \textbf{0.6585} & 0.6381 & 0.6585 \\
Aus Acc & \textbf{0.7458} & 0.7443 & 0.7346 & 0.7458\\
US Acc  & 0.7350 & \textbf{0.7613} & 0.7474 & 0.7613\\
\hline
\multicolumn{5}{c}{After PCA} \\
\hline
Twitter & \textbf{0.6487} & 0.6466 & 0.6308 & 0.6487\\          
Aus Acc & 0.7297 & 0.7133 & \textbf{0.7436} & 0.7436 \\         
US Acc  & 0.7590 & \textbf{0.7677} & 0.7495 & 0.7677 \\ 

\end{tabular}
\caption{\label{tab:cooccurrence-table} The F1 scores of our word-level GRU on each of the three datasets when the embedding layer is initialised with one of three cooccurrence embedding techniques.
  }
\end{table}

\begin{table*}[!ht]
\centering
\begin{tabular}{l | c | c | c | c | c | c}

Model architecture     & \multicolumn{5}{c|}{Bi-GRU} & Feed-forward \\ \hline
Embedding weights        & \textbf{P. dist.} & \textbf{Co-oc.} & \textbf{Word2Vec}  & \textbf{FastText} & \textbf{ELMo} & \textbf{BERT} \\ \hline

Twitter           & 0.8022 & 0.6585          & 0.7800   & 0.7323 & \textbf{0.8050} & 0.7678 \\
Aus Acc           & 0.7983 & 0.7458   & 0.7375   & 0.7601 & \textbf{0.8049} & 0.7546 \\
US Acc            & \textbf{0.8378} & 0.7677   & 0.7974   & 0.8047 & 0.8245 & 0.8241 \\
\hline
Average           & \textbf{0.8128} & 0.7240 & 0.7716   & 0.7656 & 0.8115 & 0.7822 \\


\end{tabular}
\caption{\label{tab:results-embeddings-table} The F1-Scores of our word-level LN model when initialised with each embedding technique. P. dist. and Co-oc. refer to the best-performing initialisation methods as determined in Sections~\ref{subsec:bestdist} and~\ref{subsec:cooc} respectively.
  }
\end{table*}

\begin{table*}[!ht]
\centering
\begin{tabular}{l | c | c | c | c | c | c}

Model architecture     & \multicolumn{5}{c|}{Bi-GRU} & Feed-forward \\ \hline
 Embedding weights     & \textbf{P. dist.} & \textbf{Co-oc.} & \textbf{Word2Vec}  & \textbf{FastText} & \textbf{ELMo} & \textbf{BERT} \\ \hline

Twitter           & \textbf{0.4468} & 0.2332  & 0.4274 & 0.3981 & 0.4010 & 0.3307 \\
Aus Acc           & 0.3752          & 0.1967  & 0.3173 & 0.2503 & \textbf{0.4234} & 0.2670 \\
US Acc            & 0.6818          & 0.4907   & 0.6804 & 0.7012 & \textbf{0.7289} & 0.6086 \\
\hline
Average           & 0.5013          & 0.3069  & 0.4750 & 0.4497 & \textbf{0.5178} & 0.4021 \\

\end{tabular}
\caption{\label{tab:results-embeddings-table-self} The F1-Scores of our word-level LN model when initialised with each embedding technique. The datasets do not contain the \texttt{<SELF>} label to denote words that do not require normalisation.
  }
\end{table*}

Table~\ref{tab:cooccurrence-table} shows the results of the word-level model after initialising the embeddings using the cooccurrence techniques discussed in Section~\ref{subsec:coocex}. In the lower half of the table, we use principle component analysis (PCA) to reduce the length of each embedding vector to 512.

There is little difference between the performance of each cooccurrence model. There is no clear best model across any dataset, with each model working well in certain situations. Applying PCA tended to reduce the overall F1-score slightly, except on the US Accidents dataset where it actually improved the results. However, the co-occurrence models all perform significantly worse than the randomised embeddings, indicating their inapplicability to lexical normalisation.


\subsubsection{Pre-trained embedding models}

Table~\ref{tab:results-embeddings-table} shows the results of each pre-trained embedding model. We compare these results to the best-performing probability distribution function and co-occurrence embedding method for each dataset.

Both Word2Vec and FastText tend to perform poorly on word-level lexical normalisation. Even with its character n-gram architecture, FastText struggles to be competitive with other models on the Twitter dataset. Word2Vec, on the other hand, exhibits poor results on the Australian Accidents dataset.



ELMo consistently performs the strongest out of all embedding models for the word-level models, indicating the significance of its context-dependent embeddings and purely character-level input representation. Despite taking a similarly long time to train, BERT does not perform well except on the US Accidents dataset, where it effectively ties with ELMo. This suggests BERT is not appropriate for  LN even though it performs very well on other NLP tasks.

Most notably, the probability distributions provide similar results to ELMo on every dataset, scoring slightly higher on average. The Cauchy distribution even outperforms ELMo on the US Accidents dataset. This is surprising considering the ELMo and BERT models took days to train and are considerably more advanced than a random probability distribution. 




\subsubsection{Effectiveness of the <SELF> label}
\label{subsubsec:selfeffectiveness}

To determine the effectiveness of augmenting the dataset with \texttt{<SELF>} labels (whereby a token does not require normalisation), we retrain and evaluate our models on the original datasets that do not contain these labels. These results are shown in Table~\ref{tab:results-embeddings-table-self}. The performance dropped significantly from 0.80 to 0.40 for the smaller datasets (Twitter and Australian Accidents). 

It is clear that every model performs significantly worse when the labels are not present. Pre-trained embedding models tend to outperform the randomised weights on the original datasets, except on the highly noisy Twitter dataset. ELMo typically performs best when no \texttt{<SELF>} labels are present, especially on the US Accidents dataset which contains many context-dependent tokens ("l", "r", "." and so on).

\begin{table*}[!htbp]
\centering
\begin{tabular}{l | c c | c c | c c}


& \multicolumn{2}{c|}{Model only} & \multicolumn{2}{c|}{+ Dict norm} & \multicolumn{2}{c}{\parbox{2cm}{+ Dict norm\\and flagger}}\\ \hline
\textbf{}      & \textbf{Char} & \textbf{Word}  & \textbf{Char} & \textbf{Word} & \textbf{Char} & \textbf{Word}  \\ \hline

\hline
Twitter           & 0.7848 & 0.8050 & 0.7852 & \textbf{0.8252} & 0.8102 & 0.8238 \\
Aus Acc           & 0.7432 & 0.8049 & 0.7782 & \textbf{0.8290} & 0.7936 & 0.8269 \\
US Acc            & 0.7804 & 0.8378 & 0.7827 & \textbf{0.8562} & 0.7911 & 0.8161 \\
\hline
Average           & 0.7695 & 0.8156 & 0.7820 & \textbf{0.8368} & 0.7983 & 0.8222 \\


\end{tabular}
\caption{\label{tab:results-table} The F1-Scores of each model on each of the three lexical normalisation datasets. The score of the word-level model reflects the best performing word embedding method as determined in Section~\ref{subsec:wordembeval}.
  }
\end{table*}

\subsubsection{Analysis}
\label{subsec:embanalysis}

\begin{figure*}[!ht]
\begin{minipage}{.5\linewidth}
\centering
    \captionsetup{width=.9\linewidth}
	\subfloat[Uniform distribution.] {
		\includegraphics[width=0.97\linewidth, height=140pt]{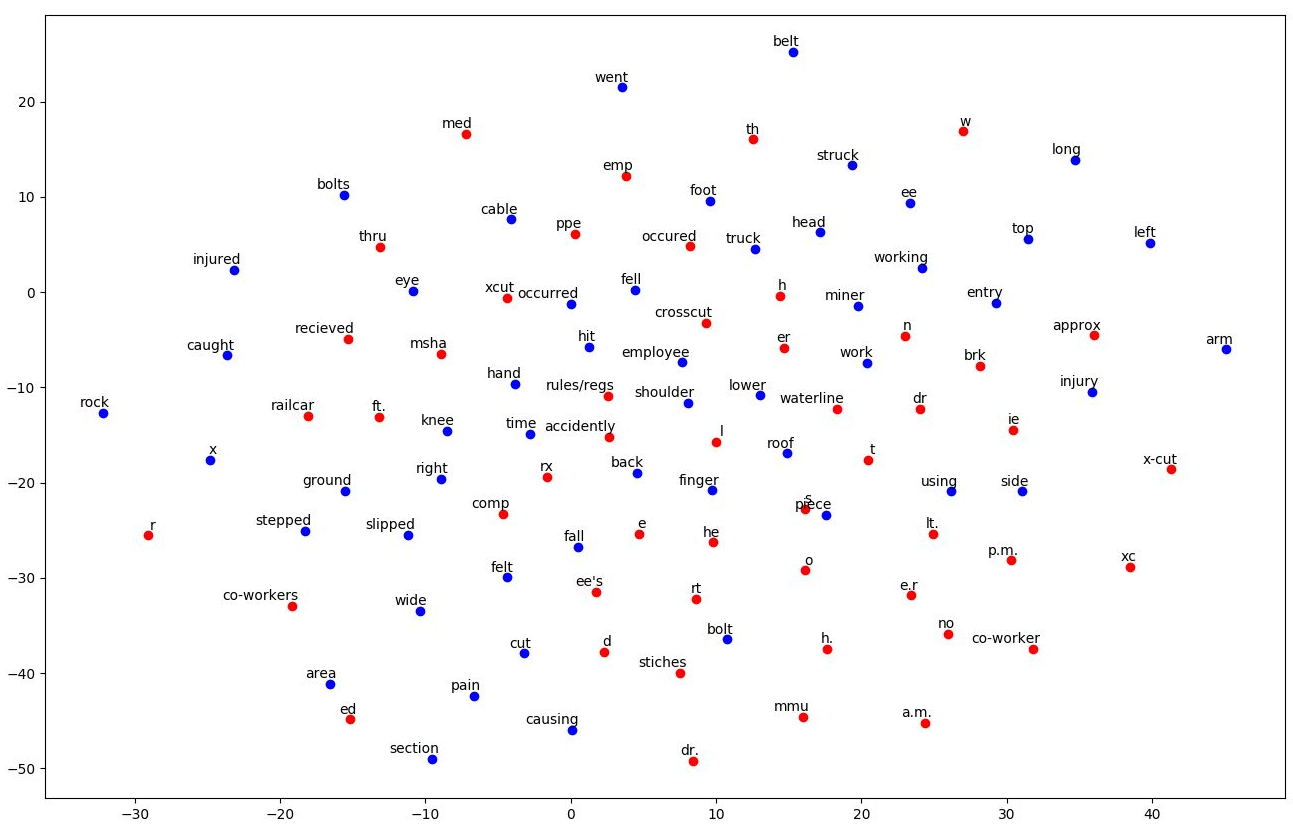}
		\label{fig:emb-uniform}
	}
\end{minipage}
\begin{minipage}{.5\linewidth}
\centering
    \captionsetup{width=.9\linewidth}
	\subfloat[Uniform distribution (after training).] {
		\includegraphics[width=0.97\linewidth, height=140pt]{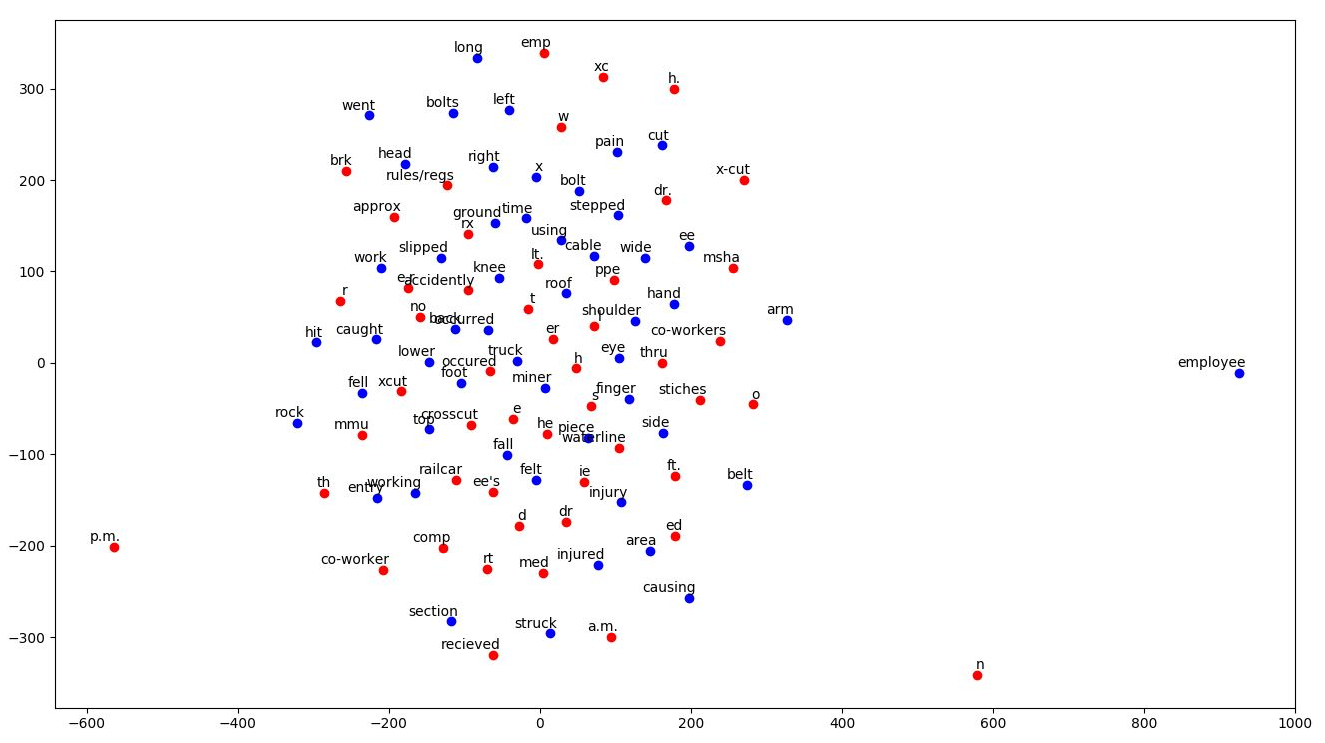}
		\label{fig:emb-uniform}
	}
\end{minipage}
\begin{minipage}{.5\linewidth}
    \captionsetup{width=.9\linewidth}
	\subfloat[Normal distribution.] {
		\includegraphics[width=0.97\linewidth, height=140pt]{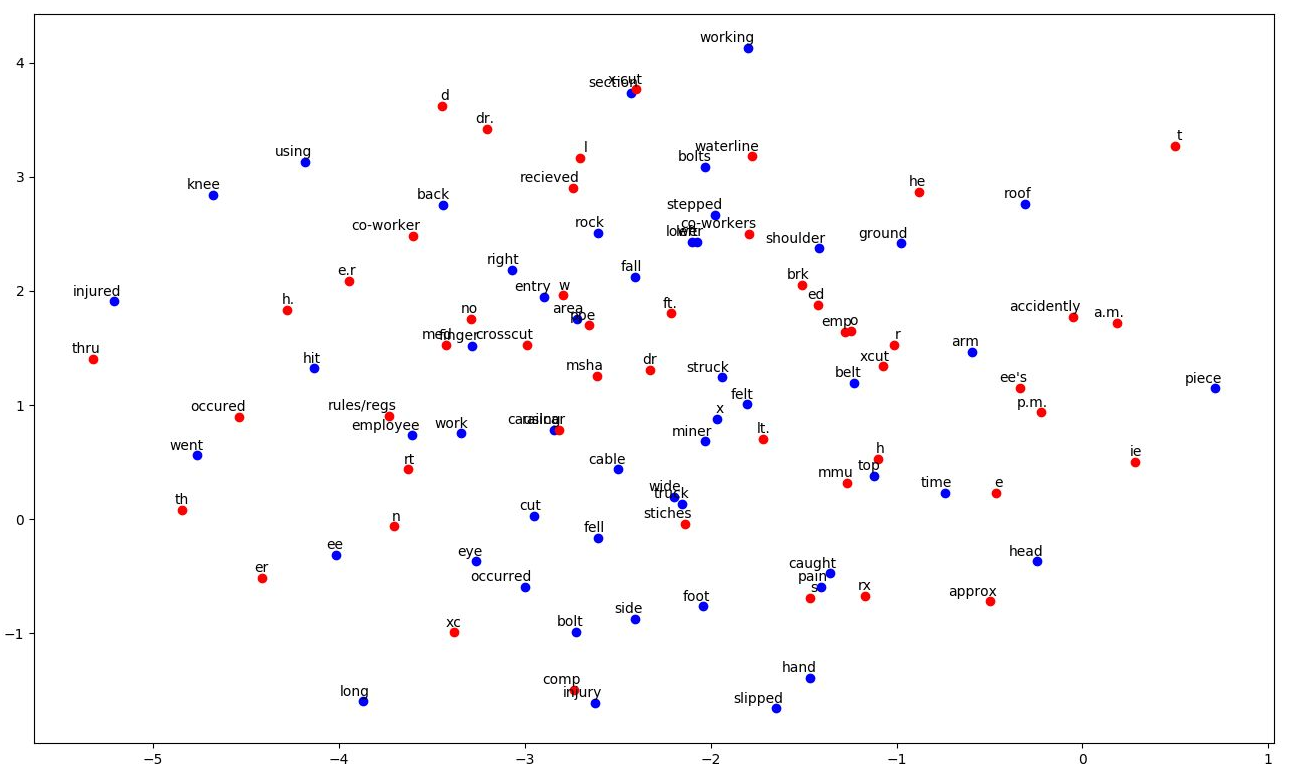}
		\label{fig:emb-normal}
	}
\end{minipage}
\begin{minipage}{.5\linewidth}
    \captionsetup{width=.9\linewidth}
	\subfloat[Normal distribution (after training).] {
		\includegraphics[width=0.97\linewidth, height=140pt]{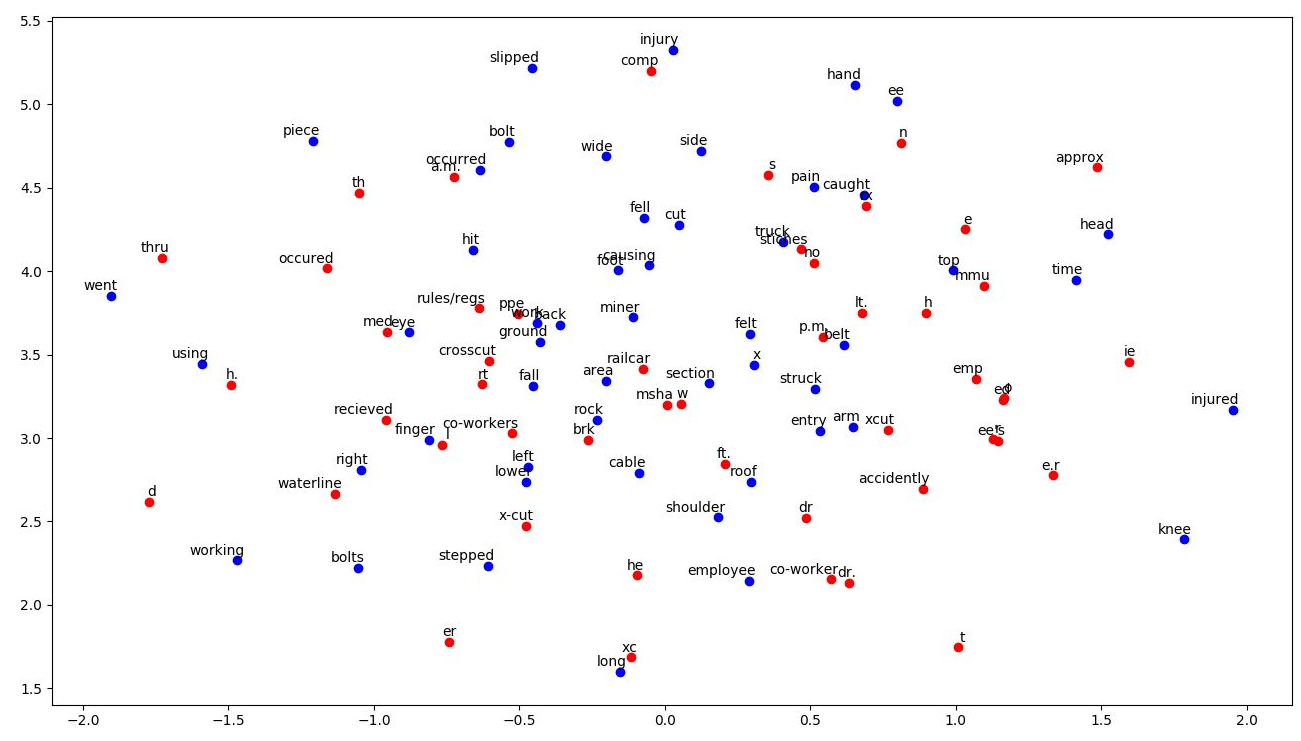}
		\label{fig:emb-normal}
	}
\end{minipage}
\begin{minipage}{.5\linewidth}
    \captionsetup{width=.9\linewidth}
	\subfloat[Cauchy distribution.] {
		\includegraphics[width=0.97\linewidth, height=140pt]{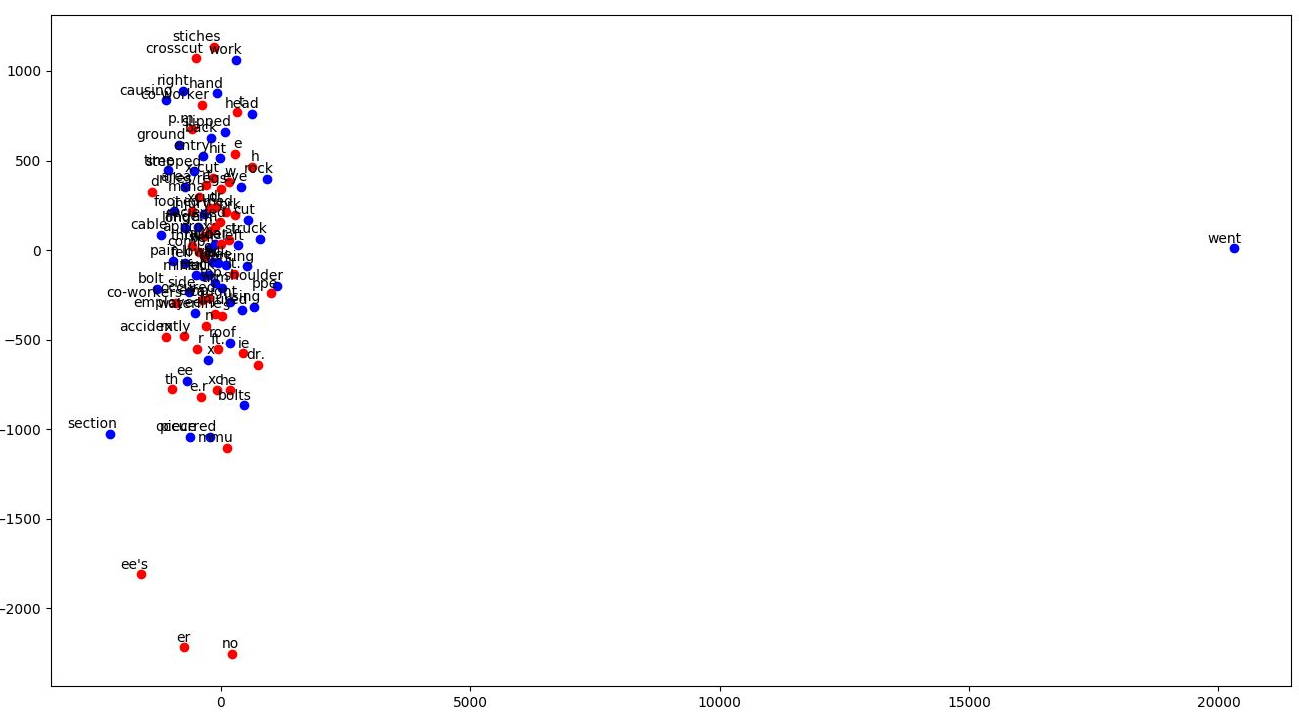}
		\label{fig:emb-cauchy}
	}
\end{minipage}
\begin{minipage}{.5\linewidth}
    \captionsetup{width=.9\linewidth}
	\subfloat[Cauchy distribution (after training).] {
		\includegraphics[width=0.97\linewidth, height=140pt]{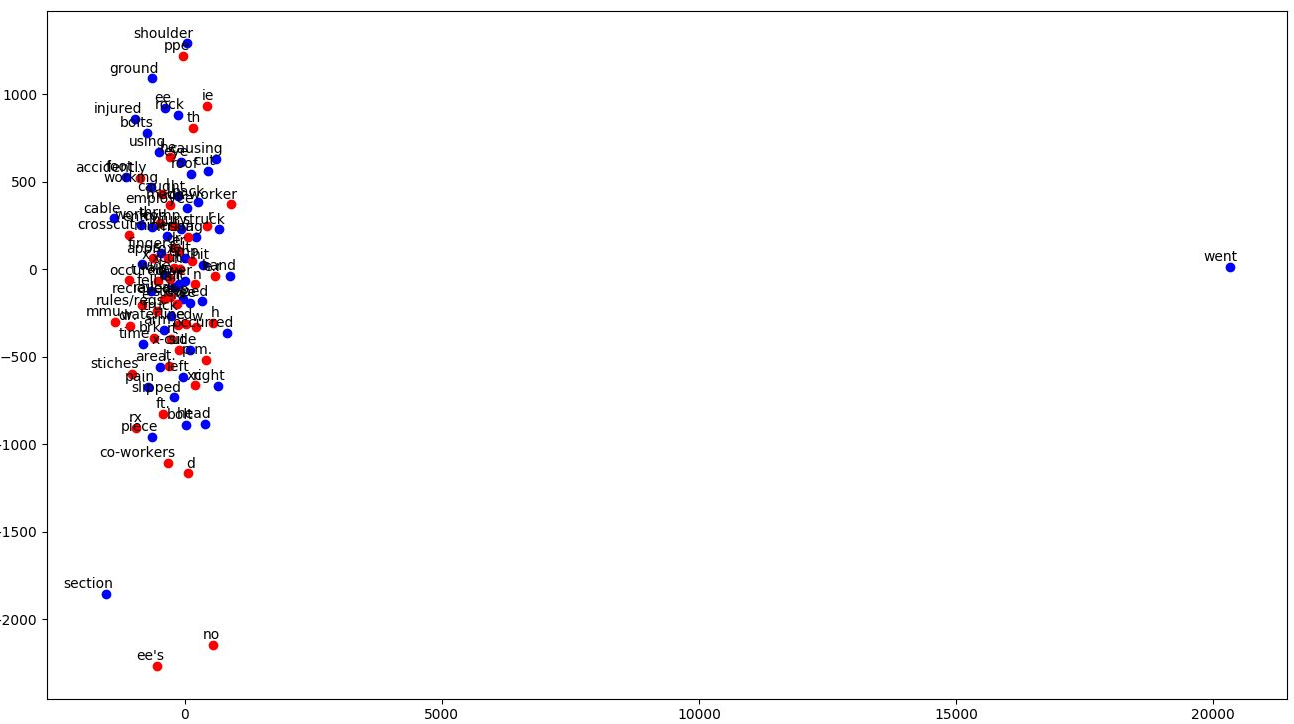}
		\label{fig:emb-cauchy}
	}
\end{minipage}
\caption{TSNE projections of the weights of the embedding layer of the word-level lexical normalisation model after being randomly initialised with a probability distribution (left) and after training (right). Correct tokens are in blue, while erroneous tokens are in red.}
	 \label{fig:emb-vis}
\end{figure*}
\begin{figure*}[!ht]
\begin{minipage}{.5\linewidth}
\centering
    \captionsetup{width=.9\linewidth}
	\subfloat[Word2Vec.] {
		\includegraphics[width=0.97\linewidth, height=140pt]{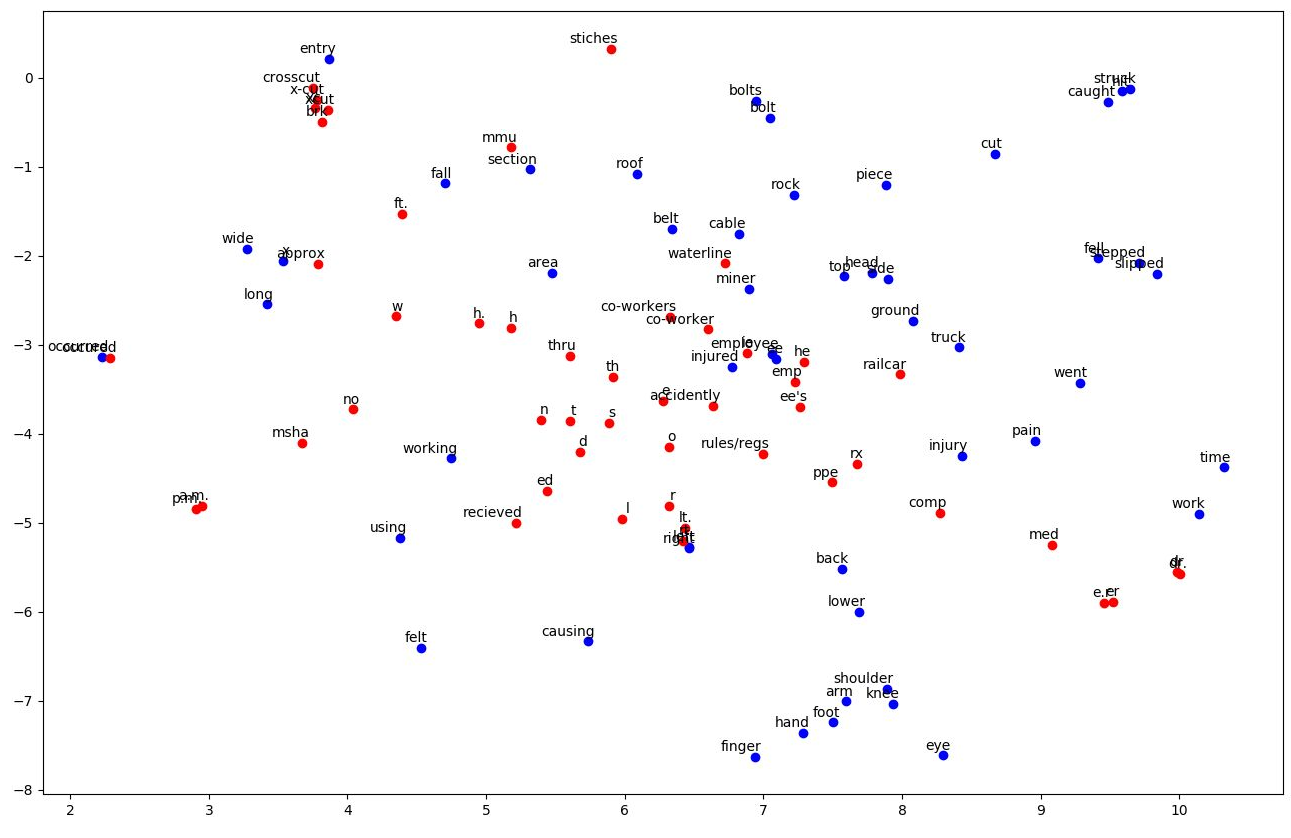}
		\label{fig:emb-word2vec}
	}
\end{minipage}
\begin{minipage}{.5\linewidth}
    \captionsetup{width=.9\linewidth}
	\subfloat[FastText.] {
		\includegraphics[width=0.97\linewidth, height=140pt]{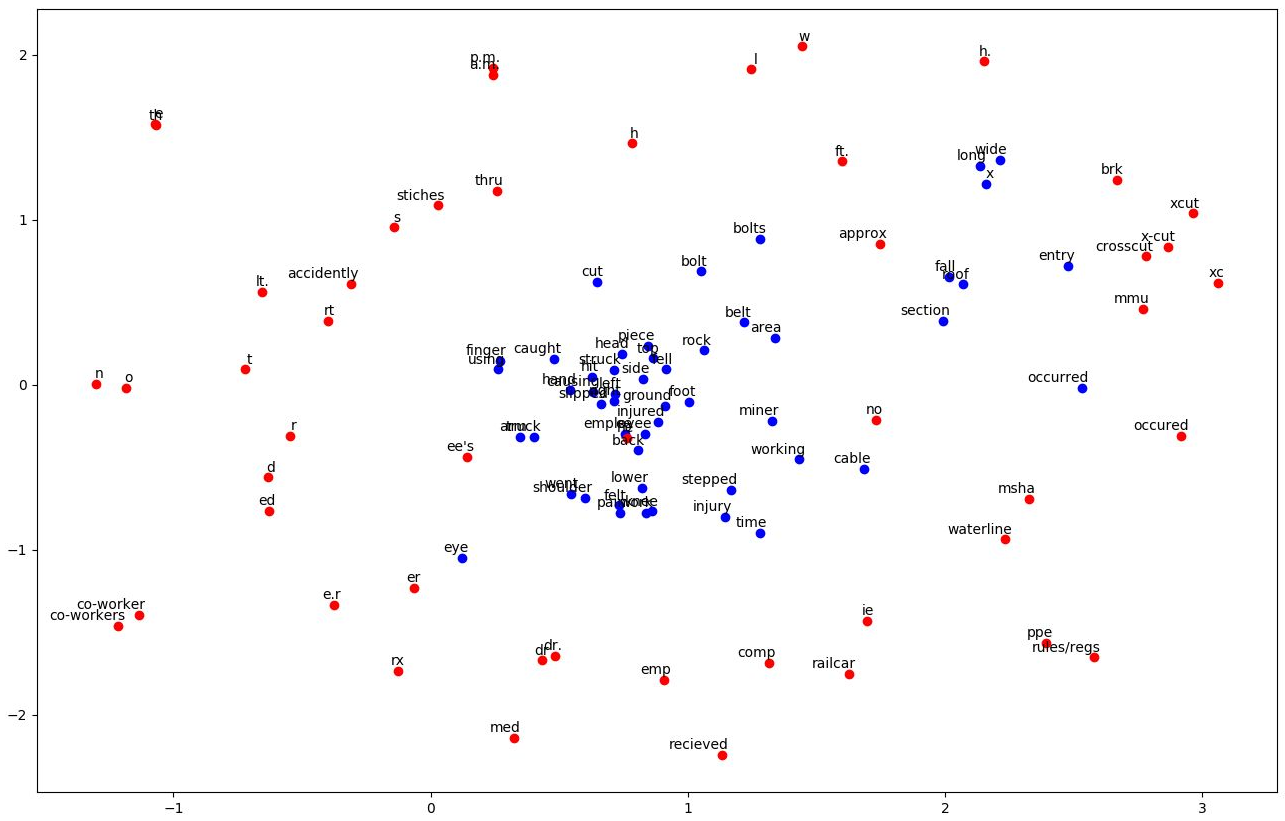}
		\label{fig:emb-fasttext}
	}
\end{minipage}
\begin{minipage}{.5\linewidth}
    \captionsetup{width=.9\linewidth}
	\subfloat[ELMo.] {
		\includegraphics[width=0.97\linewidth, height=140pt]{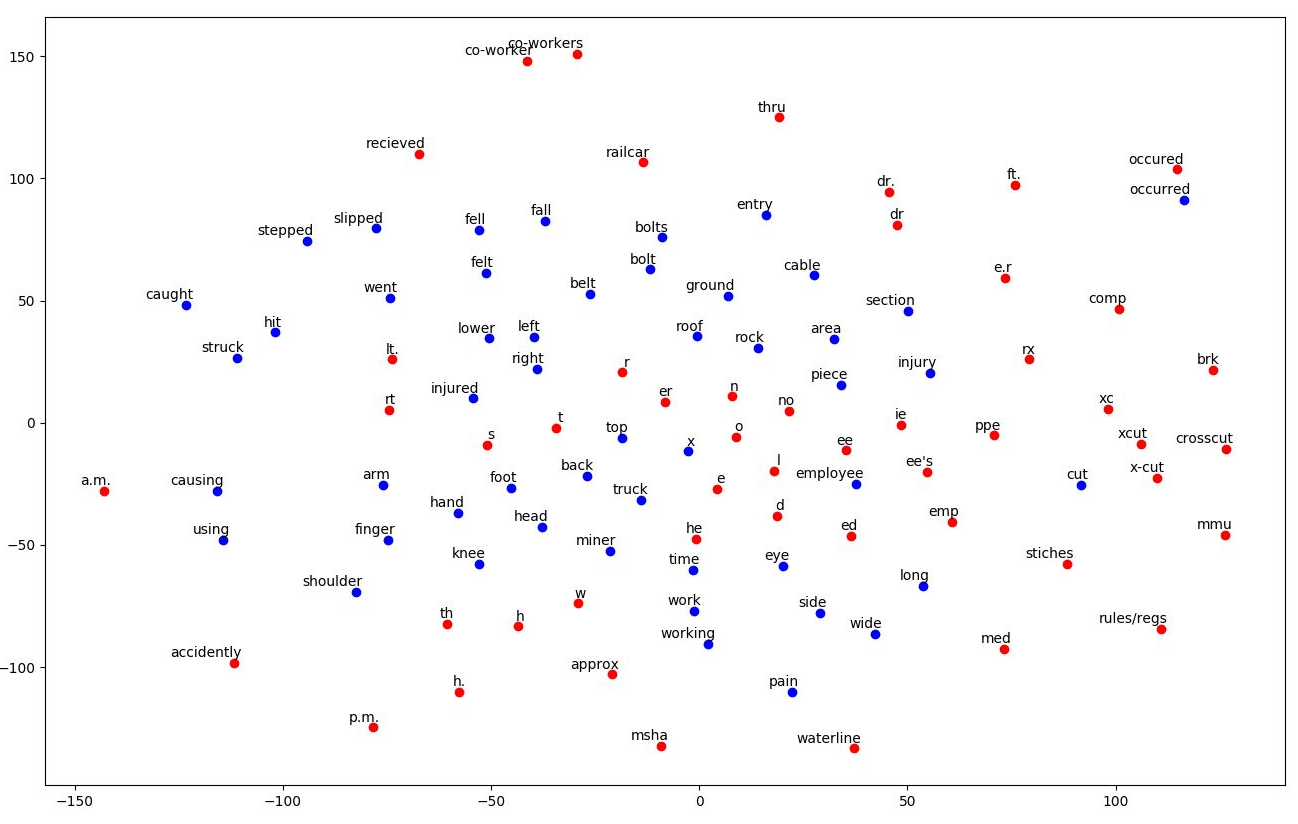}
		\label{fig:emb-elmo}
	}
\end{minipage}
\begin{minipage}{.5\linewidth}
    \captionsetup{width=.9\linewidth}
	\subfloat[BERT.] {
		\includegraphics[width=0.97\linewidth, height=140pt]{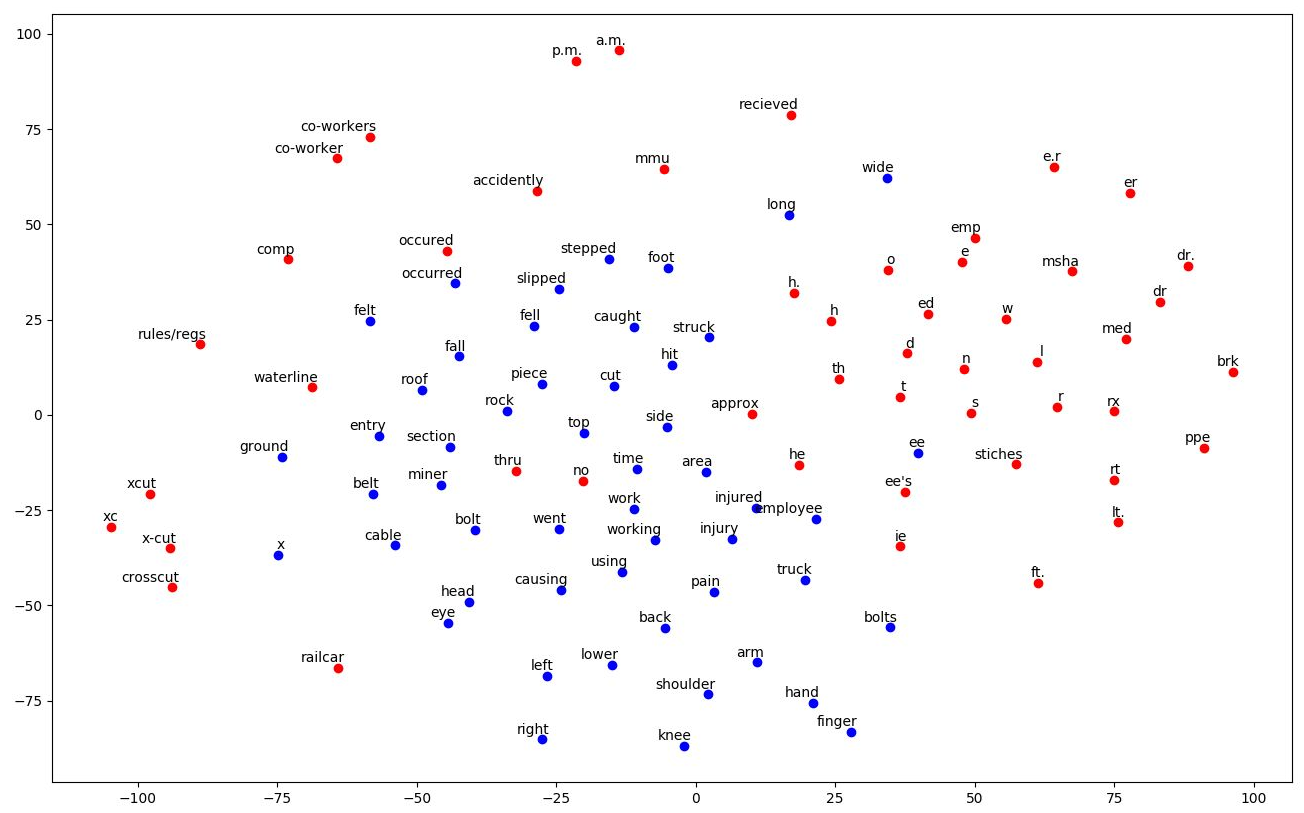}
		\label{fig:emb-bert}
	}
\end{minipage}
\caption{TSNE projections of the weights of the embedding layer of the word-level lexical normalisation model after being initialised with weights from a pre-trained embedding model.}
	 \label{fig:emb-vis-pretrained}
\end{figure*}

Figure~\ref{fig:emb-vis} displays TSNE projections of the embeddings generated via randomised embedding distributions, both before and after training.  Figure~\ref{fig:emb-vis-pretrained} displays projections of the pre-trained embedding models. For the ELMo and BERT visualisations, the embeddings of each token were taken from the average embedding of that token across all contexts. The diagrams show the 50 most common erroneous and non-erroneous tokens, coloured in red and blue respectively. 

The pretrained embedding visualisations in Figure~\ref{fig:emb-elmo} clearly show that the embeddings successfully capture contextual and character-level lexical information. This is particularly true for the ELMo embeddings, which provided the best results of all pre-trained embedding models. Erroneous words and their correctly-spelled counterparts are clustered together (e.g. "occured and occurred", "ee" and "employee"). The Cauchy embeddings, on the other hand, are undoubtedly random in nature and there is no correlation between a word's position in vector space and its meaning. The related vectors of tokens and their normalised form are also considerably further apart from one another than the ELMo vectors.

\begin{table}[!h]
\centering
\begin{tabular}{l | c | c}

              & ELMo & Cauchy \\
\hline
Total errors       & 201  & 222 \\
\hline
Norm errors   & 80   & 40 \\

\end{tabular}
\caption{ \label{tab:modelerrors} The errors made by the ELMo and Cauchy models on the US Accidents dataset.}
\end{table}

An analysis of the errors produced by the ELMo-initialised model provides clarity as to why it is outperformed by the Cauchy embeddings for the US Accidents dataset. Table~\ref{tab:modelerrors} shows that the ELMo-based model produced twice as many normalisation errors on terms present in the training set (i.e. predicting a label other than the \texttt{<SELF>} tag) compared to the Cauchy embeddings, significantly reducing precision and hence F1-score. 29 of the 80 errors made by the ELMo-based model were measurements, such as ``10'h'' being normalised to the incorrect measurement of ``8' high''. The Cauchy embeddings did not make any errors on such terms. 

The ELMo embeddings cluster all of these measurement terms together in vector space, as they are always used in similar contexts. This makes it difficult for the ELMo-based LN model to successfully disambiguate the correct term, as the input embeddings for each term are very similar. The correct form of each of these terms is wholly dependent on the term itself and is impossible to derive from context. This means that despite ELMo's ability to encapsulate context in its embeddings, this does not help with the normalisation of such terms. The embeddings provided by the Cauchy distribution, on the other hand, separate these distinct terms so that the Bi-GRU treats them as entirely distinct tokens. 

Overall we find that the most appropriate embedding technique to use for lexical normalisation depends on the dataset. Datasets containing lexically-similar terms requiring normalisation that appear in the same context, such as measurements, may see great performance from using randomised embeddings. Other datasets that do not contain such terms, e.g. Twitter, are more amenable to pre-trained embedding models because they are able to successfully encapsulate context and lexical features without adversely affecting performance.



\subsection{Word-level vs character-level lexical normalisation}

\label{subsec:wordvschar}

Our second result criterion aims to determine how our best-performing word-level LN model compares to a similar character-level LN model. Table~\ref{tab:results-table} shows the results of each model across the three datasets. The table also displays the F1-Scores after performing the dictionary normalisation and flagger techniques as explained in Section~\ref{subsec:postprocessing}.


Our best performing character-level configuration, which incorporates both the dictionary and the flagger module, slightly under-performs the results of~\citep{cox2015ncsu_sas_sam}'s Deep Encoding model despite being recurrent as opposed to feed-forward. Parameter tuning and layer-wise pretraining may help to improve the result, but it is also possible that a GRU simply does not offer an improvement over a feed-forward network on such a small dataset. 

Dictionary normalisation offers a consistent improvement to the F1-Scores, which reflects the results found by~\citep{mott2015ncsu_sas_wookhee} and ~\citep{stewart2018lexical}. The flagger module tends to work well for the character-level model, but not for the word-level model. This suggests that the word-level model is actually better at predicting whether a word requires normalisation than the flagger, which is surprising considering this is the flagger's only objective. 

The word-level model significantly outperforms both the character-level model in every case. The ability of the word-level model to utilise contextual information, combined with the robust word representations from ELMo, allow it to outshine other models despite not performing character-level normalisation like most state-of-the-art LN systems. Overall, it is clear that word-level LN has the potential to work excellently given the right embedding technique.

\subsection{Performance comparison with existing systems}

Our final result criterion compares our best performing model with state of the art social media normalisation techniques.

\begin{table}[!ht]
\centering
\begin{tabular}{lc}

\textbf{Model} & \textbf{F1-Score} \\
\hline

Random Forest                & 0.8421  \\
Lexicon + CRF + DidYouMean   & 0.8272  \\
Word-level GRU (ours)        & 0.8252  \\
Deep Contextual LSTM         & 0.8175  \\
Deep Encoding                & 0.8149  \\


\end{tabular}
\caption{\label{tab:sm-comparison-results} A performance comparison of our system against existing lexical normalisation techniques. 
  }
\end{table}

Table~\ref{tab:sm-comparison-results} shows the F-Score of our best-performing model configuration (Word-level with ELMo embeddings) on the WNUT 2015 Twitter dataset~\citep{baldwin2015shared} compared to the top four LN techniques of the competition. 

Our system outperforms all existing deep-learning based techniques on the normalisation of Twitter data. It should be noted that the existing systems listed in the table (aside from the Lexicon model) did not have access to external unlabelled Twitter data as part of the competition.

Our system does not perform as well as the two feature engineering-based systems~\citep{jin2015ncsu, supranovich2015ihs_rd}, with an F1-score difference of 0.0169 and 0.002 respectively. Our model is more general, however, as it does not require the creation of handcrafted features to work well on other domains.

%% file: section-conclusion.tex
\section{Conclusion}

We have demonstrated the effectiveness of word-level lexical normalisation by comparing the application of a number of embedding techniques. We find that randomised embeddings are capable of outperforming state-of-the-art pre-trained models, including ELMo and BERT, when the dataset contains many sparse, lexically-similar terms such as measurements. This is a surprising discovery considering these embedding models provide a significant performance boost to many other NLP tasks. It is also notable that ELMo embeddings outperform BERT embeddings in short-text LN, and that word-level models outperform character-level models in our experiments. Our word-level GRU model outperforms existing deep-learning based techniques on Twitter data. Finally, we have also released a substantial lexical normalisation dataset to the community, and hope that this dataset may be used for future LN research. 

In future, it will be useful to find a way to mitigate a key drawback of word-level lexical normalisation: the inability to normalise words to labels that have not been seen in the training data.
